\pdfoutput=1

\documentclass[11pt]{article}

\usepackage{EMNLP2023}
\usepackage{amsmath}
\usepackage{times}
\usepackage{latexsym}
\usepackage{graphicx}
\usepackage{amsmath}
\usepackage{algorithm2e}

\usepackage{makecell,tabularx}

\usepackage{hyperref}
\usepackage[T1]{fontenc}

\usepackage[utf8]{inputenc}

\usepackage{microtype}

\usepackage{inconsolata}

\title{Leveraging a Cognitive Model to Measure Subjective \\ Similarity of Human and GPT-4 Written Content}

 \author{Tailia Malloy \and Maria José Ferreira \and Fei Fang  \and Cleotilde Gonzalez \\
        tylerjmalloy@cmu.edu \ mariajor@andrew.cmu.edu \ feifang@cmu.edu \ coty@cmu.edu \\
         Carnegie Mellon University, 5000 Forbes Ave, 15222, Pittsburgh PA, USA}
  
\begin{document}
\maketitle
\begin{abstract}
Cosine similarity between two documents can be computed using token embeddings formed by Large Language Models (LLMs) such as GPT-4, and used to categorize those documents across a range of uses. However, these similarities are ultimately dependent on the corpora used to train these LLMs, and may not reflect subjective similarity of individuals or how their biases and constraints impact similarity metrics. This lack of cognitively-aware personalization of similarity metrics can be particularly problematic in educational and recommendation settings where there is a limited number of individual judgements of category or preference, and biases can be particularly relevant. To address this, we rely on an integration of an Instance-Based Learning (IBL) cognitive model with LLM embeddings to develop the Instance-Based Individualized Similarity (IBIS) metric. This similarity metric is beneficial in that it takes into account individual biases and constraints in a manner that is grounded in the cognitive mechanisms of decision making. To evaluate the IBIS metric, we also introduce a dataset of human categorizations of emails as being either dangerous (phishing) or safe (ham). This dataset is used to demonstrate the benefits of leveraging a cognitive model to measure the subjective similarity of human participants in an educational setting. 
\end{abstract} 
\section{Introduction}
When humans categorize textual information, such as when giving recommendations or learning to categorize documents, we often use our personal subjective concepts to complete the task. One example of this is giving a recommendation of a funny book to a fiend, which requires not only our own subjective conceptualization of humor, but also an understanding of the similarities and differences between ourselves and our friends. While humans perform this task with relative ease, recommendation systems \citep{ko2022survey} and educational tools \citep{nafea2019novel} typically do not have personalized measurements of subjective concepts \citep{gazdar2020new}, potentially hindering their efficacy \citep{pal2024aggregated}. 

When these systems incorporate data from human judgements to determine subjective similarity, they typically do so by pooling together as many judgements from different people as they can, and aggregate their measurement \citep{xia2015learning}. This approach relies on machine learning based methods \citep{shojaei2021mfsr}, which can be effective from a machine learning perspective, since more data can mean improved document similarity metrics on average over large datasets \citep{kusner2015word}. Focusing on individuals annotations of documents has been explored in the context of domain specific knowledge such as biomedical research papers \citep{brown2019large}, or for specific context like document summarizing \citep{zhang2003study}. 

However to date little attention has been given to the notion of individualized metrics of similarity that account for biases and constraints specifically, which are highly relevant for educational contexts \citep{chew2021cognitive}. Related to the domain of this work in particular, recent work has demonstrated a broad range of human opinions and levels of trust associated with cybersecurity concepts such as Trusted Execution Environments \citep{carreiraexplain}. This highlights the need for individualized metrics that take into account experience in training tasks such as the anti-phishing training dataset used in this work.


In this work, we propose a method for providing personalized metrics of subjective concepts that can determine the similarity between sets of text, with additional applications in selecting educational examples and providing natural language feedback. This is done by leveraging a cognitive model of human learning and decision making that can act as a digital twin to individuals, and predict their behavior and opinions on a wider set of stimuli. We focus specifically on students categorizing emails as being safe (ham) or dangerous (phishing) in a training setting to help users identify and defend against phishing email attacks. Our proposed method for providing personalized similarity metrics of documents is compared to alternative methods using a dataset of a phishing education task experiment that we additionally present in this work.  

The dataset of human annotations of emails as being either ham or phishing is described in \citep{malloy2024improving} and was made publicly available on OSF\footnote{\url{https://osf.io/wbg3r/}}. This dataset consists of human annotations of email documents that are either written by cybersecurity experts or a GPT-4 model, the emails shown to participants, and conversations between human participants and a GPT-4o model providing feedback to students. In total this dataset represents 39230 human judgements from 433 participants making decisions while observing a set from 1440 GPT-4 or human generated emails, as well as 20487 messages between human participants and the GPT-4o teacher model.

This type of learning task represents a serious challenge for traditional methods of adjusting document embedding similarity metrics to conform to human behavior, such as embedding pruning \citep{manrique2023enhancing} or embedding weighting \citep{onan2021sentiment}. This is because these approaches typically rely on a large amount of annotations collected from many participants who are expected to have the same knowledge level throughout the annotation process. Instead, we are interested in measuring the subjective similarity of documents as participants learn the document annotation task in a training setting. To do this, we employ a cognitive model that can predict the learning trajectory of each individual participant as they learn to correctly annotate these documents. 



\section{Background: Cognitive Model}
The cognitive model used in this work to predict the subjective similarity of human participants decisions on unseen emails relies on Instance Based Learning Theory (IBLT) \citep{gonzalez2003instance}. One of the benefits of employing IBL models over alternatives like Reinforcement Learning is that they base their predictions on the full history of participant experience as well as the impact that limitations like memory size and decay can have on decision making. 

IBL models have been applied onto predicting human behavior in dynamic decision making tasks, including binary choice tasks \citep{gonzalez2011instance, lejarraga2012instance}, theory of mind applications \citep{nguyen2022theory}, and practical applications such as identifying phishing emails \citep{cranford2019modeling, malloy2024applying}, cyber defense \citep{cranford2020toward}, and cyber attack decision-making \citep{aggarwal2022designing}.  


\subsection{Activation}
IBL models work by storing instances $i$ in memory $\mathcal{M}$, composed of utility outcomes $u_i$ and options $k$ composed of features $j$ in the set of features $\mathcal{F}$ of environmental decision alternatives. These options are observed in an order represented by the time step $t$, and the time step that an instance occurred in is given $\mathcal{T}(i)$. Option values are determined by selecting the action that maximizes the blended value $\mathcal{V}_k(t)$. In calculating this activation, the similarity between instances in memory and the current instance is represented by summing over all attributes the value $S_{ij}$, which is the similarity of attribute $j$ of instance $i$ to the current state. This gives the activation equation as: 
\begin{equation}
\begin{split}
A_i(t) = \ln \Bigg( \sum_{t' \in \mathcal{T}_i(t)} (t - t')^{-d}\Bigg) + \\
\mu \sum_{j \in \mathcal{F}} \omega_j (S_{ij} - 1) + \sigma \xi
\label{eq:activation}
\end{split}
\end{equation}
The parameters that are set either by modelers or set to default values are the decay parameter $d$; the mismatch penalty $\mu$; the attribute weight of each $j$ feature $\omega_j$; and the noise parameter $\sigma$. The default values for these parameters are $(d,\mu,\omega_j,\sigma) = (0.5, 1, 1, 0.25)$. The value $\xi$ is drawn from a normal distribution $\mathcal{N}(-1,1)$ and multiplied by the noise parameter $\sigma$ to add random noise to the activation.    

\subsection{Similarity Measure}
The definition of the similarity measure $S_{ij}$ is highly influential in the behavior of the IBL model, as it determines which instances from memory are drawn from to predict utility. In simple binary choice tasks without attributes \citep{gonzalez2011instance, lejarraga2012instance}, the similarity metric can be defined as the equality function $S_{ij} = 1$ if $i == j $ else $0$. In more complex domains such as the phishing email identification task used in this work, one approach is to use the embeddings of emails to compare the similarity of instances, and rely on the cosine similarity metric to compute the similarity of instances in memory \citep{malloy2024applying}. The model presented in this work relies on an initial baseline similarity metric, the standard cosine similarity, to then build more individual specific metrics of similarity. 

\subsection{Probability of Retrieval}
The probability of retrieval represents the probability that a single instance in memory will be retrieved when estimating the value associated with an option. To calculate this probability of retrieval, IBL models apply a weighted soft-max function onto the memory instance activation values $A_i(t)$ giving the equation:
\begin{equation}
P_{i}(t) = \dfrac{\exp{A_i(t)/\tau}}{\sum_{i' \in \mathcal{M}_k}\exp{A_{i'}(t)/\tau}}
\label{eq:retrieval}
\end{equation}
The parameter that is either set by modelers or set to its default value is the temperature parameter $\tau$, which controls the uniformity of the probability distribution defined by this soft-max equation. The default value for this parameter is $\tau = \sigma \sqrt{2}$.

\subsection{Blended Value}
The blended value determines the ultimate action selected by the model and is calculated of an option $k$ at time step $t$ according to the utility outcomes $u_i$ weighted by the probability of retrieval of that instance $P_i$ and summing over all instances in memory $\mathcal{M}_k$ to give the equation:
\begin{equation}
V_k(t) = \sum_{i \in \mathcal{M}_k} P_i(t)u_i
\label{eq:blending}
\end{equation}

These blended values are used to determine the action $a_{t+1}$ selected by the model at the next time step.  
\begin{equation}
a_{t+1} = \max_{k \in K} V_k(t)
\label{eq:blending}
\end{equation}
In standard IBL models, this action can be used in simulations to allow the model to gain experience in a given task. In model tracing, which is used in the method proposed in this work, the memory of instances is made up of the past observations and decisions of the participant, with the action representing a prediction of their future behavior.

\section{Phishing Email Categorization Dataset}
The first component of this dataset is human behavioral experiment data from a study on human categorization of emails. This experiment compared human document annotation when categorizing emails as phishing (dangerous) or ham (safe). The conditions of this experiment varied depending on the email author (Human or GPT-4) and style (plain-text or GPT-4 stylized). There was also a comparison of the method of selecting emails to show to participants, either randomly selected, or chosen using an IBL model (IBL or Random). Finally, we compared the type of feedback given to participants between positive and negative point feedback and a natural language conversation with an GPT-4o chat-bot (Points or Written). 

This experiment included 10 pre-training trials without feedback, 40 training trials with feedback, and 10 post-training trials without feedback. During all trials, participants made judgments of emails as phishing or ham and indicated their confidence in their judgment as well as which action out of 6 possibilities they would select after receiving the email. We recruited 433 participants online through the Amazon Mechanical Turk (AMT) platform. Participants (150 Female, 280 Male, 3 Non-binary) had an average age of 40.3 with a standard deviation of 11.02 years.  Participants were compensated with a base payment of \$3-5 with the potential to earn up to a \$12-15 bonus payment depending on performance and the length of the experiment. This experiment was approved by the Carnegie Mellon University Institutional Review Board, and the study was pre-registered on OSF.

The second component of this dataset is the emails shown to participants, which were either written by human cybersecurity experts, a GPT-4 model working alone, or a combination of human and GPT-4 model work. 360 base emails written by human experts were used to form three additional versions of these base emails. These alternative versions included a `human-written gpt4-styled' version that used the email body written by human experts, the `gpt4-written and gpt4-styled' version that was fully rewritten by GPT-4, and the `gpt4-written plaintext-styled' version that stripped the HTML and CSS styling applied by the GPT-4 model. These emails as well as the original prompts to generate them are included in dataset on OSF. 



\section{Methods of Measuring Similarity}
\subsection{Human Subjective Similarity}
LLM embeddings have been suggested as a method of measuring human similarity judgements \citep{bhatia2022cognitive}, while also capturing the wide range of individuals similarity measures. Additionally, comparisons of LLM behavior have also demonstrated human-like variability \citep{bhatia2024exploring}, suggesting these embeddings could be useful for capturing the variety of human similarity judgements. Cognitive models that rely on representations of information from GAI models have been shown to adequately account for the wide range of human behavior \citep{Mitsopoulos2023Psychologically}.  

However, for these methods to function properly there must be a connection between the way that similarity is measured in humans and GAI models. Previous applications in applying visual GAI models onto representing decision-making tasks in humans relied on the close connection to these model representations and human representations \citep{higgins2016beta, higgins2021unsupervised}. For this reason, we devised a metric of human subjective similarity that takes into account the confidence of document categorization as well as the time it takes participants to categorize documents. 

\begin{figure}[t] 
\begin{centering}
  \includegraphics[width=7cm]{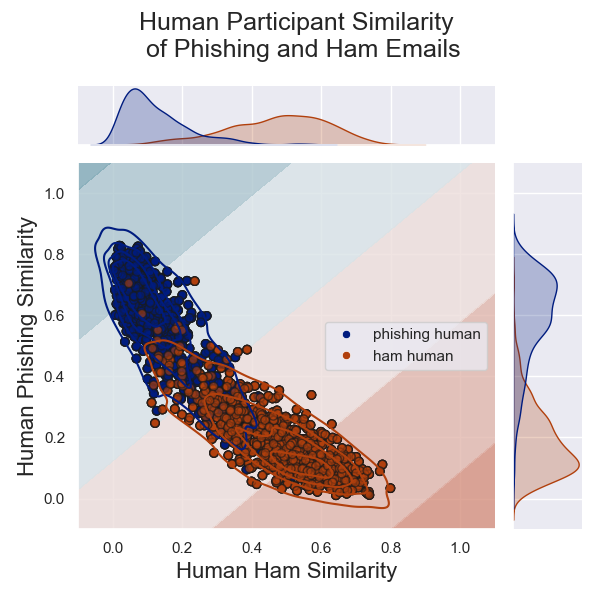} 
  \caption{Human participant similarity measure for all 1440 phishing (blue) and ham (orange) emails. Shaded region is a logistic regression.}\label{fig:Figure0}
 \end{centering} 
\end{figure}

To determine the human subjective similarity measure, we use the category of human participant annotations, their annotation confidence, and the speed of their annotation. For accuracy and confidence, a higher value in our human subjective similarity metric signifies that participants were more likely to categorize an emails as being a member of that group, and more confident in their categorization. For reaction time, a lower value indicates that the document is more immediately obviously a member of a group and thus has a higher similarity to other members of that group. The result is a value that is difficult for a standard similarity metric to account for, as the annotations made in this dataset occurred in a learning setting where earlier trials had less accuracy, which also impacted reaction time and confidence.  

The reaction time and confidence weighted subjective similarity of an email $x$ is given by multiplying the probability of a human participant categorizing that email as category $c$ giving $cs(x|c) = p(c|x)r(c|x)c(c|x)$. where $p(c|x)$ is the probability of categorization, $r(c|x)$ is the reaction time normalized to between 0 and 1, and $c(c|x)$ is the confidence additionally normalized to between 0 and 1. The soft-max of this $cs(x|c)$ value is the resulting similarity metric, with the equation shown in the supplementary materials\footnote{\url{https://osf.io/wbg3r/}}.

\begin{equation}
HS(x,x') = \dfrac{cs(x|c)cs(x'|c)}{\sum_{c' \in C} cs(x|c) \sum_{c' \in C} cs(x'|c')} 
\label{eq:humanSubjective}
\end{equation}

Figure \ref{fig:Figure0} shows the average human similarity measures for each of the 1440 emails in the dataset. The human ham and human phishing similarities are calculated according to Equation \ref{eq:humanSubjective} by averaging the accuracy, reaction time, and confidence across all participants in the dataset. It is also possible to calculate this subjective similarity for an individual only using the documents that subject categorized. We next compare similarity measures in their ability to capture individual human subjective similarity.

\begin{figure}[t] 
\begin{centering}
  \includegraphics[width=7cm]{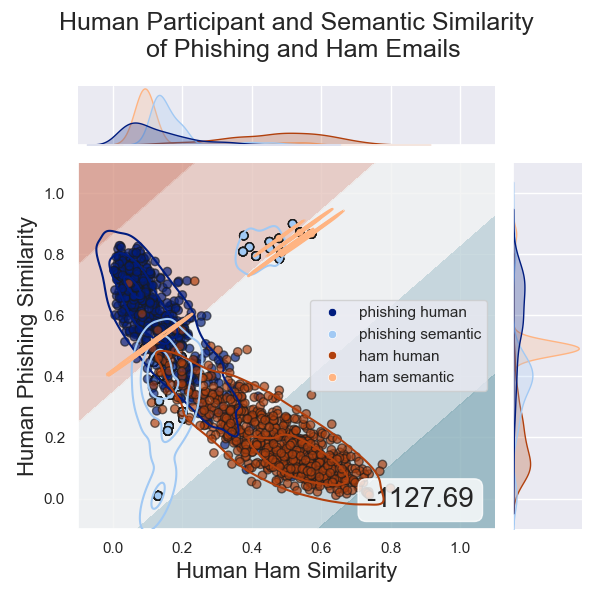} 
  \caption{Semantic and human participant similarity for phishing (blue) and ham (orange) emails. Shaded region is a logistic regression. The Kernel Density Estimate log probability score between each distribution is shown on the bottom right, higher is better.}\label{fig:Figure1}
 \end{centering} 
\end{figure}

\begin{figure}[t] 
\begin{centering}
  \includegraphics[width=7cm]{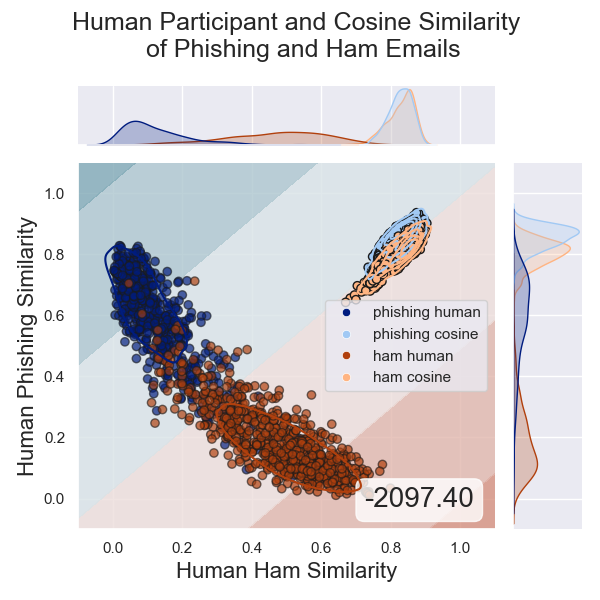} 
  \caption{Cosine and human participant similarity for phishing (blue) and ham (orange) emails. Shaded region is a logistic regression. The Kernel Density Estimate log probability score between each distribution is shown on the bottom right, higher is better.}\label{fig:Figure2}
 \end{centering} 
\end{figure}

\subsection{Semantic Similarity}
One method of measuring the similarity of documents is to employ semantic information contained in documents and compare the similarities and differences between documents in terms of these semantic categories. This has been done in the past in applications such as  topic modeling \citep{rehurek_lrec}, document annotation \citep{pech2017semantic}, and calculating document similarity \citep{qurashi2020document}.

In this dataset, semantic similarity can be calculated using the categorizations of email features that were originally made by the cybersecurity experts who created the base email dataset. These features are Link Mismatch, Offer, Urgent, Subject Suspicious, Request Credentials, and Sender Mismatch. Figure \ref{fig:Figure1} plots these semantic similarity measures for each of the 1440 emails in our dataset, and compares the distribution of these similarities to our human subjective similarity metric. 

These semantic similarity metrics are close to human similarity for phishing emails (blue), but highly diverge from the similarity scores of ham emails (orange). This results in a low Kernel Density Estimate log probability score (-1127.69) between the two distributions compared to the semantic similarity metric. This metric compares the likelihood that the data-points in the human similarity metric distribution would have come from the semantic similarity distribution, summing all log probabilities. This low score is due to the fact that the majority of ham emails are very sparse for all of the six semantic categories previously mentioned. 


\begin{figure}[t] 
\begin{centering}
  \includegraphics[width=7cm]{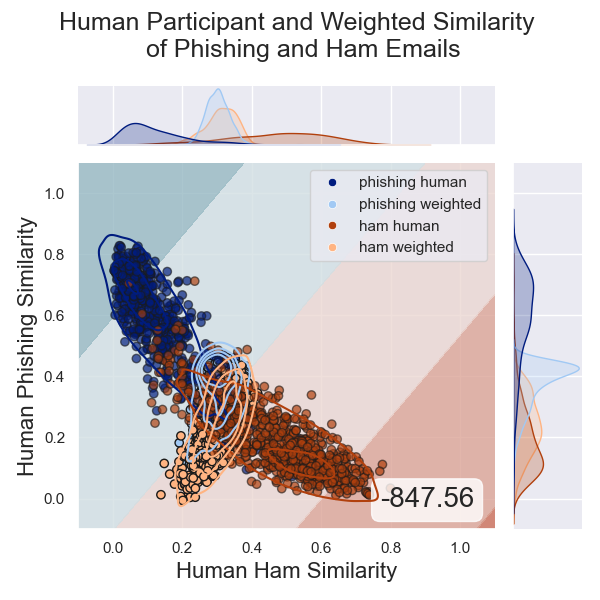} 
  \caption{Cosine and human participant similarity for phishing (blue) and ham (orange) emails. Shaded region is a logistic regression. The Kernel Density Estimate log probability score between each distribution is shown on the bottom right, higher is better.}\label{fig:Figure3}
 \end{centering} 
\end{figure}

\subsection{Cosine Similarity}
Cosine similarity is the most commonly used metric of similarity of word and document embeddings, with many applications from classification \citep{park2020methodology}, recommendation systems \citep{khatter2021movie}, educational tutorial systems \citep{wu2023matching}, question answering \citep{aithal2021automatic}, and more \citep{patil2023survey}. However, there are limitations to using cosine similarity such as in documents with high-frequency words \citep{zhou2022problems}, and the presence of false information \citep{borges2019combining}, both of which are concerns for phishing email education.  

The cosine similarity metric is calculated using an embedding of size 3072 formed by the `text-embedding-3-large' model, accessed through the OpenAI API, these document embeddings are additionally included in our presented dataset. The cosine similarity of each email embedding is compared to the mean embedding of that category and shown in Figure \ref{fig:Figure2}, and compared to our metric of human subjective similarity. From this, we can see that on average the embeddings are calculated as being significantly more similar to each other compared to the subjective similarities of human participants. This results in a lower Kernel Density Estimate log probability score (-2097.40) between the two distributions compared to the semantic similarity metric. 

\subsection{Weighted Cosine Similarity}
Distance weighted cosine similarity is a common method employed in utilizing embeddings \citep{li2013distance}, which has been applied onto measuring similarity of online instruction in educational settings \citep{lahitani2016cosine}, as well as several cybersecurity specific applications like ransomware detection \citep{moussaileb2021survey}, and inside attacker detection \citep{khan2019malicious}. In this work, we employ weighted cosine similarities of embeddings formed from emails categorized as being either ham or phishing, and compare it to human subjective similarity judgements. This weighting is done by learning a weight transformation of size 3072, the same as the embedding size, which is applied onto the embedding prior to calculating the similarity. The results of this weighting are shown in Figure \ref{fig:Figure3}, which compares the average human participant subjective similarity and the weighted cosine similarity of email embeddings.

The KDE log probability score between weighted cosine similarities of phishing and ham emails compared to human subjective similarity has increased to -847.56 from the unweighted KDE score of -2097.40, surpassing the semantic similarity score at -1127.69. These improved similarity metrics indicate that weighting cosine similarity based on data from a large dataset of human participants can result in a metric that more accurately reflects the average of human subjects' subjective similarity metrics. 

\subsection{Pruning Document Embeddings}
\begin{figure}[t] 
\begin{centering}
  \includegraphics[width=7cm]{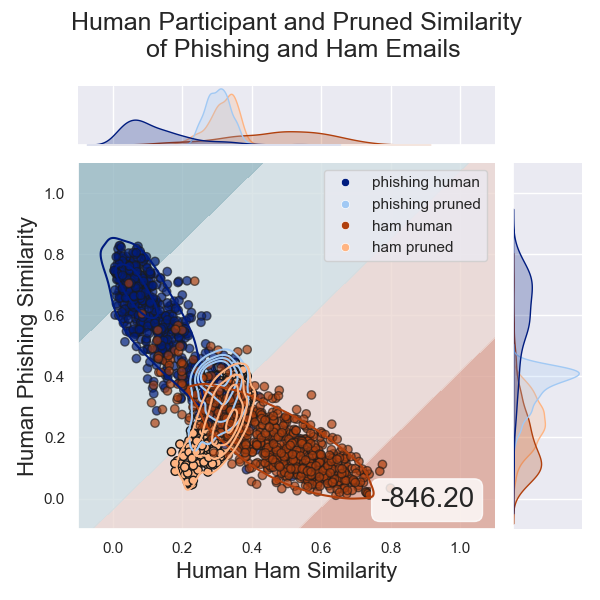} 
  \caption{Pruned cosine and human participant similarity for phishing (blue) and ham (orange) emails. Shaded region is a logistic regression. The Kernel Density Estimate log probability score between each distribution is shown on the bottom right, higher is better.}\label{fig:Figure4}
 \end{centering} 
\end{figure}
Another method of comparison documents is embedding pruning, where embeddings are reduced in size based on feedback from human categorizations to better account for their subjective similarity \citep{manrique2023enhancing}. These approaches function by reducing the number of embedding values that are used in comparison, and are similar to the weighting method except with 0 or 1 values. We structured our embedding pruning method to select only the top 500 embedding values, representing just under 20\% of the size of the embedding, as was done in \citep{manrique2023enhancing}. These top predictive embedding values are retained, while all other values are masked to 0. After this, cosine similarity can be calculated with the standard approach, resulting in the similarity shown in Figure \ref{fig:Figure5}. Compared to the weighted cosine similarity method, the pruned cosine similarity has roughly the same KDE log probability score. 

\section{Ensemble Similarity}
\begin{figure}[t] 
\begin{centering}
  \includegraphics[width=7cm]{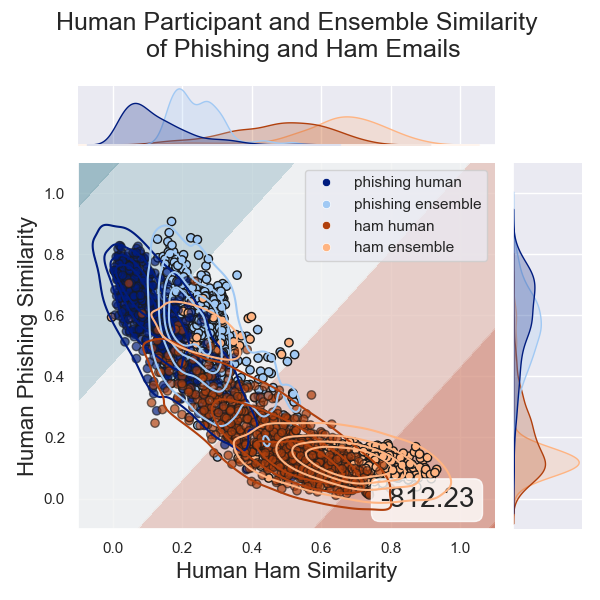} 
  \caption{Ensemble and human participant similarity for phishing (blue) and ham (orange) emails. Shaded region is a logistic regression. The Kernel Density Estimate log probability score between each distribution is shown on the bottom right, higher is better.}\label{fig:Figure5}
 \end{centering} 
\end{figure}
The final comparison method is based on using an ensemble of each of the previous similarity metrics, weighted to maximize the similarity to the average of the human subjective similarity metrics. This approach has been applied to document matching for patent documents \citep{yu2024semantic}, which requires the similarity of document embeddings be calculated to determine a match. This ensemble approach has the highest KDE log probability score of any individual method by itself, at a value of -812.23. Looking at the KDE distributions above and to the right of the scatter plot in \ref{fig:Figure5} demonstrates the high similarity of the ensemble similarity metric (light blue and light orange) and the human participant similarity metric (blue and orange). While this method is effective at resulting in a similarity metric that closely matches the average over all participants, it still does not fit as well to individual participants, as will be shown in our proposed model. 

\section{Instance-Based Individualized Similarity (IBIS)}
\begin{figure}[t] 
\begin{centering}
  \includegraphics[width=7cm]{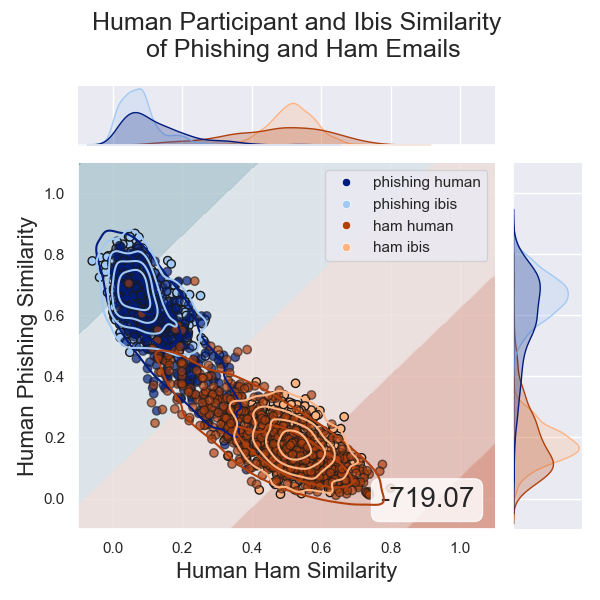} 
  \caption{IBIS and human participant similarity for phishing (blue) and ham (orange) emails. Shaded region is a logistic regression. The Kernel Density Estimate log probability score between each distribution is shown on the bottom right, higher is better.}\label{fig:Figure6}
 \end{centering} 
\end{figure}
To determine an individual participant's metric of similarity, we employ an IBL model that is serving as a digital twin of the participant. The result in an Instance-Based Individualized Similarity (IBIS) metric. The benefits of IBIS are in the ability to predict human judgements on unseen documents or feedback from recommendations, and enhance measurements of subjective similarity. Importantly, these predictions of human behavior are not merely relying on a separate machine learning based technique, but rather a cognitive model that is inspired by the human cognitive mechanisms underlying decision making and thus able to account for natural biases and constraints in humans. 

Predictions of Instance-Bases Individual Similarity are done using an IBL model that is currently serving as a digital twin with the same experience as an individual participant. Using this we determine the value that the IBL model assigns to predicting a category $c$ as $V_k(c|x)$, or the value the IBL model assigns to choosing option $c$ as the category of document $x$. Then, we can divide this value by the same categorization value assigned to each alternative categorization of the same document. This results in the IBIS metric which can be calculated after each decision is made by a participant, pseudo-code for the IBIS algorithm, The code-base for the IBIS method including all comparison methods, data, and scripts to generate similarity measures and figures is made available\footnote{\href{https://github.com/TylerJamesMalloy/cognitive-similarity}{github.com/TylerJamesMalloy/cognitive-similarity}}.

\section{Case Study of IBIS: Individuals in Phishing Email Education Dataset}
\begin{figure*}[t] 
\begin{centering}
  \includegraphics[width=16cm]{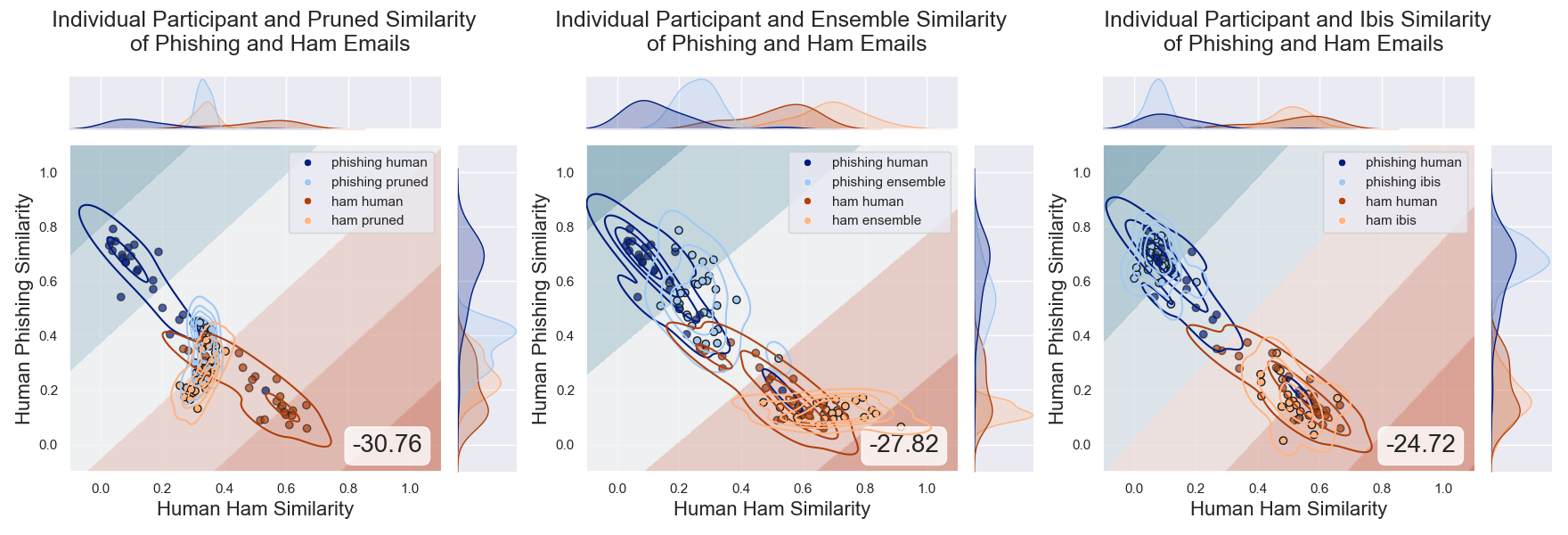} 
  \caption{Top performing similarity metrics and individual participant similarity for phishing and ham emails. Shaded region is a logistic regression. The lower value is the individual KDE score}\label{fig:Figure7}
 \end{centering} 
\end{figure*}
Previous comparisons of similarity metrics and human participant behavior compared the average of human performance. To highlight the benefits of the IBIS method, we replicate these calculations with one individual from the experiment. Here, the individual similarity of phishing and ham emails is based only on a single individuals categorization, confidence, and reaction time in their judgement. These graphs are shown for illustration in Figure \ref{fig:Figure7}, with the average accuracy of logistic regression of similarity metrics predicting individual participant similarity metrics reported in table \ref{table}. 

The KDE score of the similarities for the pruned cosine method is -30.76, and for the ensemble method it is -27.82. Note that these scores are much lower than the entire dataset scores since they are calculated using only the emails observed by the participant. Meanwhile, the IBIS metric gives a KDE score of -24.72. From this we can see that the IBIS method effectively learns the similarity measures of individual participants. These results are used for illustrative purposes, and the averages across all participants for regression accuracy, as well as the DKE score for individuals, is presented in Table \ref{table}.

An important aspect of individual similarity comparisons of the IBIS method is that it can compare emails that were not originally presented to an individual, meaning there are more embedding similarities used in the logistic regression and KDE score calculation. This comparison demonstrates the benefits of using a cognitively inspired method of modeling human participant decisions making that takes into account biases and cognitive constraints. 

The results is a prediction of behavior that can accurately fill in the gaps of unseen elements of the dataset that have not been observed by a participant. This method more accurately predicts the subjective similarity of participants. Importantly, this is done while initially limiting the cognitive model to observing a single decision made by these participants, and increasing this data as the participant makes more decisions. This is important for the functioning of the IBL model as using too many instances in memory can slow compute performance. 

The final comparison shown in the right most columns of Table \ref{table} shows the percent accuracy in using the previously described logistic regressions, shown on all figure results, in predicting the categorization of participants based on the similarity metric applied onto the emails they observed. This regression has the potential to predict the annotations of individuals, similarly to the IBL model. Comparing these measures shows that the best performance comes from the IBIS metric when predicting participant annotations.

\begin{table*}[t!]
\begin{center}
\begin{tabular}{||c c c c||} 
 \hline
 Similarity  & KDE Score & KDE Score  & Regression \\ 
 Metric &  Average Participants & Individuals  &  Accuracy \\ 
 \hline\hline
 Semantic Similarity \citep{qurashi2020document} & -1127.69 & -37.69$\pm$1.19 & 0.46$\pm$0.11\\
 \hline
 Cosine Similarity \citep{park2020methodology} & -2097.40 &  -47.26$\pm$2.27 & 0.52$\pm$0.10 \\ 
 \hline
 Embedding Weighting \citep{onan2021sentiment} & -847.56 &  -29.28$\pm$2.32 & 0.86$\pm$0.14 \\
 \hline
 Embedding Pruning \citep{manrique2023enhancing} & -846.20 & -30.39$\pm$2.76 & 0.86$\pm$0.04 \\
 \hline
 Ensemble Similarity \citep{yu2024semantic} & -812.23 &  -28.64$\pm$3.28 & 0.89$\pm$0.12 \\
 \hline
 IBIS (proposed)
 & \textbf{-719.07} &  \textbf{-23.17$\pm$3.29}  & \textbf{0.93$\pm$0.04} \\ [1ex] 
 \hline
\end{tabular}
\caption{Comparison of the six previously described methods in their similarity to human behavior. Similarity to average participants is performed across the entire dataset of human judgements (see Figures 1-6). Similarity to individuals and regression accuracy are both done for each individual participant (see Figure 7). For all values higher is better. Reported values are means of all participants measured individually $\pm$ standard deviations.}
\label{table}
\end{center}
\end{table*} 

\section{Discussion}
Many applications of LLMs are interested in tailoring use cases to individuals, even when little information is known about that individual. While many approaches of individualization exist but have typically relied on advanced machine learning techniques. The method proposed in this work is relatively simple from a mathematical perspective, though there is a strength in its reliance on theories of cognition that underlie human learning and decision making. The result is a simple to understand and easy to implement method of calculating similarities of unseen documents using a cognitive model, which can augment datasets that contain only a small number of decisions. 

The general method described here, of augmenting subjective similarity metrics with predicted decisions from a cognitive model, could be applied onto various other scenarios. This includes settings that leverage representations formed of visual information such as $\beta$-Variational Autoencoders \citep{higgins2016beta}, which have been related to biological representation formation \citep{higgins2021unsupervised}. Overall, we believe that this method is useful for any application where the experience of end-users impacts future decisions.

For instance, in visual learning settings VAEs have been integrated with cognitive models to predict human utility learning of abstract visual information \citep{malloy2024efficient}. Other integrations of Generative AI into cognitive models includes use of LLMs as a knowledge repositories within cognitive models \citep{kirk2023exploiting}. In particular, ConceptNet \citep{speer2017conceptnet} has previously been integrated into a cognitive model for question answering \citep{huet2021cacda}. Future research should investigate how additional uses of LLMs in integrations of cognitive models can aid in educational settings. 

Overall, the results in this work demonstrate the usefulness of cognitive models in serving as digital twins to human participants. Leveraging these models and integrating their results into Large Language Model techniques has the potential to make measurements from these models more cognitively grounded. While there are existing methods of incorporating human behavior through the use of large datasets collected from many participants, these do not necessarily account for biases and constraints. The method proposed in this work takes these features of human learning and decision making into account in developing a similarity metric.

\section{Limitations}
The semantic similarity metric suffered from the sparsity of semantic categories in ham emails, additional annotations could raise the performance of this metric and can be explored in future work. However, this ensemble method was partially responsible for the high KDE score of the ensemble method, as it allowed for an integration of both semantic information and embedding similarity. Our IBIS method still outperformed the ensemble method suggesting that this ensemble alone does not address the issues of alternative methods. 

One limitation inherent in IBL cognitive models is the time requirements to compare the current instance to all instances in memory. This may make the proposed model unsuitable for applications that rely on large datasets of individual behavior. However, methods in instance compression exist for IBL models \citep{nguyen2023speedyibl}. In this setting, we were able to predict individual participant's decisions fast enough that this was unnecessary.

The specific application we investigated is somewhat unique in that it is based on training human participants to make categorization judgements of textual information of one of two categories. Additionally, the task of annotating whether an email is phishing or ham relies heavily on a small number of features within the email. Namely, if an email contains a link that redirects to a nefarious website, or requests personal information, then it should be labelled as phishing. While students rely on many queues to make their judgements, the annotation is in reality simple. Future work in the area of learning subjective similarity metrics should expand into more complex domains.

\section{Acknowledgement}
This research was sponsored by the Army Research Office and accomplished under Australia-US MURI Grant Number W911NF-20-S-000, and the AI Research Institutes Program funded by the National Science Foundation under AI Institute for Societal Decision Making (AI-SDM), Award No. 2229881. Compute resources and GPT model credits were provided by the Microsoft Accelerate Foundation Models Research Program grant ``Personalized Education with Foundation Models via Cognitive Modeling".

\clearpage 
\newpage
\bibliography{anthology,custom}
\bibliographystyle{acl_natbib}

\end{document}